# Forecasting Economic and Financial Time Series: ARIMA vs. LSTM

**Sima Siami Namin[1], Akbar Siami Namin[2]**

[1.] Graduate Research Assistant and Ph.D. Student, Department of Agricultural and Applied Economics

*sima.siami-namini@ttu.edu*

[2.] Associate Professor, Department of Computer Science

*akbar.namin@ttu.edu*

Texas Tech University

Lubbock, TX, USA

March 15, 2018

**Abstract**

Forecasting time series data is an important subject in economics, business, and finance. Traditionally, there are several techniques to effectively forecast the next lag of time series data such as univariate Autoregressive (AR), univariate Moving Average (MA), Simple Exponential Smoothing (SES), and more notably Autoregressive Integrated Moving Average (ARIMA) with its many variations. In particular, ARIMA model has demonstrated its outperformance in precision and accuracy of predicting the next lags of time series. With the recent advancement in computational power of computers and more importantly developing more advanced machine learning algorithms and approaches such as deep learning, new algorithms are developed to forecast time series data. The research question investigated in this article is that whether and how the newly developed deep learning-based algorithms for forecasting time series data, such as "*Long Short-Term Memory (LSTM)*", are superior to the traditional algorithms. The empirical studies conducted and reported in this article show that deep learning-based algorithms such as LSTM outperform traditional-based algorithms such as ARIMA model. More specifically, the average reduction in error rates obtained by LSTM is between 84 - 87 percent when compared to ARIMA indicating the superiority of LSTM to ARIMA. Furthermore, it was noticed that the number of training times, known as "*epoch*" in deep learning, has no effect on the performance of the trained forecast model and it exhibits a truly random behavior.

# FORECASTING ECONOMIC AND FINANCIAL TIME SERIES: ARIMA VS. LSTM

## 1 Introduction

Prediction of economic and financial time series data is a challenging task mainly due to the unprecedented changes in economic trends and conditions in one hand and incomplete information on the other hand. Market volatility in recent years has produced serious issues for economic and financial time series forecasting. Therefore, assessing the accuracy of forecasts is necessary when employing various forms of forecasting methods, and more specifically forecasting using regression analysis as they have many limitations in applications.

The main objective of this article is to investigate which forecasting methods offer best predictions with respect to lower forecast errors and higher accuracy of forecasts. In this regard, there are varieties of stochastic models in time series forecasting. The most well known method is univariate "*Auto-Regressive Moving Average (ARMA)*" for a single time series data in which Auto-Regressive (AR) and Moving Average (MA) models are combined. Univariate "*Auto-Regressive Integrated Moving Average (ARIMA)*" is a special type of ARMA where differencing is taken into account in the model. Multivariate ARIMA models and Vector Auto-Regression (VAR) models are the other most popular forecasting models, which in turn, generalize the univariate ARIMA models and univariate autoregressive (AR) model by allowing for more than one evolving variable.

Machine learning techniques and more importantly deep learning algorithms have introduced new approaches to prediction problems where the relationships between variables are modeled in deep and layered hierarchy. Machine learning-based techniques such as Support Vector Machines (SVM) and Random Forests (RF) and deep-learning based algorithms such as Recurrent Neural Network (RNN), and Long Short-Term Memory (LSTM) have gained lots of attentions in recent years with their applications in many disciplines including finance. Deep learning methods are capable of identifying structure and pattern of data such as non-linearity and complexity in time series forecasting. In particular, LSTM has been used in many application domains such as natural language processing [Tarwani and Edem, 2017], handwriting recognition [Graves et al. 2009], speech recognition [Robinson, 2002; Eyben 2009; Graves et al. 2013; Sak et al. 2014], time-series prediction [Hochreiter and Schmidhuber, 1997; Gers, 2000; Yim, 2002; Zhang, 2005; Graves et al., 2009, 2013; Schmidhuber, 2015; Brownlee, 2016; Gamboa, 2017; Roondiwala, 2017] as well as its applications in economics and finance data such as predicting the volatility of the S&P 500 [Kohzadi, 1996; Giles at al. 2001; Huck, 2009; Xiong, 2015] and measuring the impact of incorporating news for selected companies [Siah and Myers, Joshi et al. 2013; Ding et al. 2015].

An interesting and important research question is then the accuracy and precision of traditional forecasting techniques when compared to deep learning-based forecasting algorithms. To the best of our knowledge, there is no specific empirical evidence for using LSTM method in forecasting economic and financial timer series data to assess its performance and compare it with traditional econometric forecasting methods such as ARIMA.

This paper compares ARIMA and LSTM models with respect to their performance in reducing error rates. As a representative of traditional forecast modeling, ARIMA is chosen due to the non-stationary property of the data collected and modeled. As a representative of deep learning-based algorithms, LSTM method is used due to its use in preserving and training the features of data for a longer period of time. The paper provides an in-depth guidance on data processing and training of LSTM models for a set of economic and financial time series data. The contribution of this paper to the literature is to forecast a variety of economic and financial time series by using ARIMA and LSTM models. The key contributions of this paper are:

- Conduct an empirical study and analysis with the goal of investigating the performance of traditional forecasting techniques and deep learning-based algorithms.
- Compare the performance of LSTM and ARIMA with respect to minimization achieved in the error rates in prediction. The study shows that LSTM outperforms ARIMA. The average reduction in error rates obtained by LSTM is between 84 - 87 percent when compared to ARIMA indicating the superiority of LSTM to ARIMA.
- Investigate the influence of the number of training times. The study shows that the number of training performed on data, known as "*epoch*" in deep learning, has no effect on the performance of the trained forecast model and it exhibits a truly random behavior.

The article is organized as follows. Section 2 outlines the state of the art of time series forecasting. Section 3 discusses the mathematical background of ARIMA and LSTM. Section 4 describes an experimental study for ARIMA versus LSTM model. The ARIMA and LSTM algorithms developed and compared are presented in Section 5. The results of data analysis and empirical results are presented in Section 6. Section 7 discusses the impact of the number of iterations on fitting model. Finally, Section 8 concludes the paper.

## 2   Time Series Forecasting: The State-of-the-Art

Time series analysis and dynamic modeling is a research interesting area with a great number of applications in business, economics, finance and computer science. The aim of time series analysis is to study the path observations of time series and build a model to describe the structure of data and predict the future values of time series. Due to the importance of time series forecasting in many branches of applied sciences, it is essential to build an effective model with the aim of improving the forecasting accuracy. A variety of the time series forecasting models have been evolved in the literature.

Time series forecasting is traditionally performed in econometrics using ARIMA models which is generalized by Box and Jenkins [1970]. ARIMA has been a standard method for time series forecasting for a long time. Even though ARIMA models are very pervalent in modeling economical and financial time series [Banerjee 2005; Khashei and Bijari, 2011; Alonso and García-Martos, 2012; Adebiyi et al. 2014], they have some major limitations [Armstrong 2001; Earnest et al. 2005;]. For istance, in a simple ARIMA model, it is hard to model the nonlinear relationships between variables. Furthermore, it is assumed that there is a constant standard deviation in errors in ARIMA model, whch in practice it may not be satisfied. When an ARIMA model is integrated with a Generalized Autoregressive Conditional Heteroskedasticity (GARCH) model, this assumption can be relaxed. On the other hand, the optimization of an GARCH model and its parameters might be challenging and problematic [Kane, 2014].

Recently, new techniques in deep learning have been developed to address the challenges related to the forecasting models. LSTM (Long Short-Term Memory) is a special case of Recurrent Neural Network (RNN) method that was initially introduced by Hochreiter and Schmidhuber [Hochreiter and Schmidhuber, 1979]. Even though it is a relatively new approach to address prediction problems, deep learning-based approaches have gained popularities among researchers. For instance, Krauss et al. [Krauss et al , 2017] use various forms of forecasting models such as deep learning, gradient-boosted trees, and random forests to model S&P 500 constitutes. Surprisingly, they reported that deep learning-based modeling underperfomed gradient-boosted trees and random forests. Additionally, Krauss et al. report that training neural networks and consequently deep leanring-based algorithms is very difficult. Lee and Yoo [Lee and Yoo, 2017] introduced an RNN-based approach to predict stock returns. The idea was to build portfolios by adjusting the threhshold levels of return by internal layers of the RNN built. In this article, we compare the performance of an ARIMA model with the LSTM model in the prediction of economics and financial time series to deermine the optimal qualities of involved variables in a typical prediction model.

## 3 Mathematical Background

This section reviews the mathematical background of the time series techniques used and compared in this article. More specifically, the background knowledge of Autoregressive Integrated Moving Average (ARIMA) and the deep learning-based technique, Long Short-Term Memory (LSTM) is presented.

### 3.1 Autoregressive Integrated Moving Average Model (ARIMA)

ARIMA [Pesaran, 2015] is a generalized model of Autoregressive Moving Average (ARMA) that combines Autoregressive (AR) process and Moving Average (MA) processes and builds a composite model of the time series. As acronym indicates, $ARIMA\ (p, d, q)$ captures the key elements of the model:

- **AR**: *Autoregression*. A regression model that uses the dependencies between an observation and a number of lagged observations ($p$).
- **I**: *Integrated*. To make the time series stationary by measuring the differences of observations at different time ($d$).
- **MA**: *Moving Average*. An approach that takes into accounts the dependency between observations and the residual error terms when a moving average model is used to the lagged observations ($q$).

A simple form of an AR model of order $p$, i.e., AR ($p$), can be written as a linear process given by:

$$x_t = c + \sum_{i=1}^{p} \phi_i\, x_{t-i} + \varepsilon_t$$

Where $x_t$ is the stationary variable, $c$ is constant, the terms in $\phi_i$ are autocorrelation coefficients at lags $1, 2, \dots, p$ and $\varepsilon_t$, the residuals, are the Gaussian white noise series with mean zero and variance $\sigma_\varepsilon^2$. An MA model of order $q$, i.e., MA ($q$), can be written in the form:

$$x_t = \mu + \sum_{i=0}^{q} \theta_i \varepsilon_{t-i}$$

Where $\mu$ is the expectation of $x_t$ (usually assumed equal to zero), the $\theta_i$ terms are the weights applied to the current and prior values of a stochastic term in the time series, and $\theta_0 = 1$. We assume that $\varepsilon_t$ is a Gaussian white noise series with mean zero and variance $\sigma_\varepsilon^2$. We can combine these two models by adding them together and form an ARMA model of order $(p, q)$:

$$x_t = c + \sum_{i=1}^{p} \emptyset_i \, x_{t-i} + \varepsilon_t + \sum_{i=0}^{q} \theta_i \, \varepsilon_{t-i}$$

Where $\emptyset_i \neq 0, \theta_i \neq 0$, and $\sigma_\varepsilon^2 > 0$. The parameters $p$ and $q$ are called the AR and MA orders, respectively. ARIMA forecasting, also known as Box and Jenkins forecasting, is capable of dealing with non-stationary time series data because of its "*integrate*" step. In fact, the "*integrate*" component involves differencing the time series to convert a non-stationary time series into a stationary. The general form of a ARIMA model is denoted as $ARIMA\ (p, d, q)$.

With seasonal time series data, it is likely that short run non-seasonal components contribute to the model. Therefore, we need to estimate seasonal $ARIMA$ model, which incorporates both non-seasonal and seasonal factors in a multiplicative model. The general form of a seasonal ARIMA model is denoted as $ARIMA\ (p, d, q) \times (P, D, Q)S$, where $p$ is the non-seasonal AR order, d is the non-seasonal differencing, q is the non-seasonal MA order, P is the seasonal AR order, D is the seasonal differencing, Q is the seasonal MA order, and S is the time span of repeating seasonal pattern, respectively. The most important step in estimating seasonal ARIMA model is to identify the values of $(p, d, q)$ and $(P, D, Q)$. Based on the time plot of the data, if for instance, the variance grows with time, we should use variance-stabilizing transformations and differencing. Then, using autocorrelation function (ACF) to measure the amount of linear dependence between observations in a time series that are separated by a lag $p$, and the partial autocorrelation function (PACF) to determine how many autoregressive terms $q$ are necessary and inverse autocorrelation function (IACF) for detecting over differencing, we can identify the preliminary values of autoregressive order $p$, the order of differencing $d$, the moving average order $q$ and their corresponding seasonal parameters $P$, $D$ and $Q$. The parameter $d$ is the order of difference frequency changing from non-stationary time series to stationary time series.

### 3.2 Long Short-Term Memory (LSTM)

LSTM [Patterson, 2017] is a kind of Recurrent Neural Network (RNN) with the capability of remembering the values from earlier stages for the purpose of future use. Before delving into LSTM, it is necessary to have a glimpse of what a neural network looks like.

#### *3.2.1 Artificial Neural Network (ANN)*

A neural network consists of at least three layers namely: 1) an input layer, 2) hidden layers, and 3) an output layer. The number of features of the dataset determines the dimensionality or the number of nodes in the input layer. These nodes are connected through links called "*synapses*" to the nodes created in the hidden layer(s). The synapses links carry some weights for every node in the input layer. The weights basically play the role of a decision maker to decide which signal, or input, may pass through and which may not. The weights also show the strength or extent to the hidden layer. A neural network basically learns by adjusting the weight for each synopsis.

In the hidden layers, the nodes apply an activation function (e.g., sigmoid or tangent hyperbolic (*tanh*)) on the weighted sum of inputs to transform the inputs to the outputs, or predicted values. The output layer generates a vector of probabilities for the various outputs and selects the one with minimum error rate or cost, i.e., minimizing the differences between expected and predicted values, also known as the cost, using a function called *SoftMax*.

The assignments to the weights vector and thus the errors obtained through the network training for the first time might not be the best. To find the most optimal values for errors, the errors are "*back propagated*" into the network from the output layer towards the hidden layers and as a result the weights are adjusted. The procedure is repeated, i.e., epochs, several times with the same observations and the weights are re-adjusted until there is an improvement in the predicted values and subsequently in the cost. When the cost function is minimized, the model is trained.

### 3.2.2   *Recurrent Neural Network (RNN)*

A recurrent neural network (RNN) is a special case of neural network where the objective is to predict the next step in the sequence of observations with respect to the previous steps observed in the sequence. In fact, the idea behind RNNs is to make use of sequential observations and learn from the earlier stages to forecast future trends. As a result, the earlier stages data need to be remembered when guessing the next steps. In RNNs, the hidden layers act as internal storage for storing the information captured in earlier stages of reading sequential data. RNNs are called "*recurrent"* because they perform the same task for every element of the sequence, with the characteristic of utilizing information captured earlier to predict future unseen sequential data. The major challenge with a typical generic RNN is that these networks remember only a few earlier steps in the sequence and thus are not suitable to remembering longer sequences of data. This challenging problem is solved using the "*memory line*" introduced in the Long Short-Term Memory (LSTM) recurrent network.

### 3.2.3   *Long Short-Term Memory (LSTM)*

LSTM is a special kind of RNNs with additional features to memorize the sequence of data. The memorization of the earlier trend of the data is possible through some gates along with a memory line incorporated in a typical LSTM. The internal structure of an LSTM cell is demonstrated in Diagram 1:

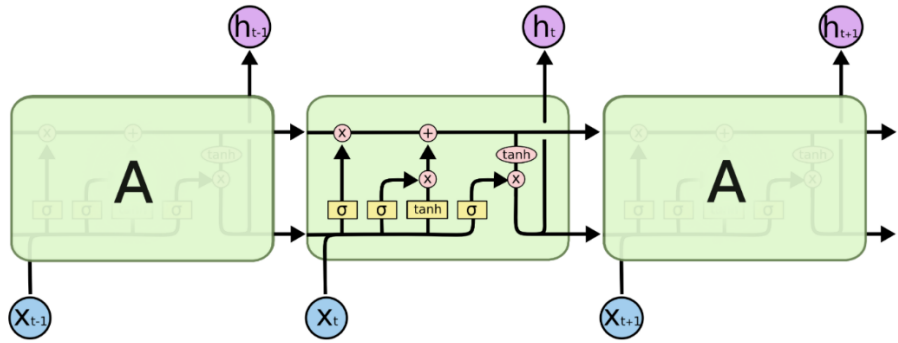

**Diagram 1. The internal structure of an LSTM [Colah's Blog, 2015].**

LSTM is a special kind of RNNs with additional features to memorize the sequence of data. Each LSTM is a set of *cells*, or system modules, where the data streams are captured and stored. The cells resemble a transport line (the upper line in each cell) that connects out of one module to another one conveying data from past and gathering them for the present one. Due to the use of some gates in each cell, data in each cell can be disposed, filtered, or added for the next cells. Hence, the gates, which are based on sigmoidal neural network layer, enable the cells to optionally let data pass through or disposed.

Each sigmoid layer yields numbers in the range of zero and one, depicting the amount of every segment of data ought to be let through in each cell. More precisely, an estimation of zero value implies that "*let nothing pass through*"; whereas; an estimation of one indicates that "*let everything pass through*." Three types of gates are involved in each LSTM with the goal of controlling the state of each cell:

- **Forget Gate** outputs a number between 0 and 1, where 1 shows "*completely keep this*"; whereas, 0 implies "*completely ignore this.*"
- **Memory Gate** chooses which *new* data need to be stored in the cell. First, a sigmoid layer, called the "*input door layer*" chooses which values will be modified. Next, a *tanh* layer makes a vector of new candidate values that could be added to the state.
- **Output Gate** decides what will be yield out of each cell. The yielded value will be based on the cell state along with the filtered and newly added data.

## 4 ARIMA vs. LSTM: An Experimental Study

With the goal of comparing the performance of ARIMA and LSTM, the authors conducted a series of experiments on some selected economic and financial time series data. The main research questions investigated through this work are as follows:

1. **RQ1**. Which algorithm, ARIMA or LSTM, performs more accurate prediction of time series data?
2. **RQ2**. Does the number of training times in deep learning-based algorithms influence the accuracy of the trained model?

### 4.1 Data Sets

The authors extracted historical monthly financial time series from Jan 1985 to Aug 2018 from the Yahoo finance Website[1]. The monthly data included Nikkei 225 index (N225), NASDAQ composite index (IXIC), Hang Seng Index (HIS), S&P 500 commodity price index (GSPC), and Dow Jones industrial average index (DJ). Moreover, the authors also collected monthly economics time series for different time periods from the Federal Reserve Bank of St. Louis[2], and the International Monetary Fund (IMF) Website[3]. The data included Medical care commodities for all urban consumers, Index 1982-1984=100 for the period of Jan 1967 to July 2017 (MC), Housing for all urban consumers, Index 1982-1984=100 for the period of Jan 1967 to July 2017 (HO), the trade-weighted U.S. dollar index in terms of major currencies, Index Mar 1973=100 for the period of Aug 1967 to July 2017 (EX), Food and Beverages for all urban consumers, Index 1982-1984=100 for the period of Jan 1967 to July 2017 (FB), M1 Money Stock, billions of dollars for the period of Jan 1959 to July 2017 (MS), and Transportation for all urban consumers, Index 1982-1984=100 for the period of Jan 1947 to July 2017 (TR).

### 4.2 Data Preparation

Each financial time series data set features a number of variables: Open, High, Low, Close, Adjusted Close and Volume. The authors chose the "Adjusted Close" variable as the only feature of financial time series to be fed into the ARIMA and LSTM models. Each economic and financial time series data set was split into two subsets: training and test datasets where 70% of each dataset was used for training and the remaining 30% of each dataset was used for testing the accuracy of models. Table 1 lists the number of time series observations for each dataset.

**Table 1. The number of time series observations.**

| Stock | # Observations | | Total | Stock | # Observations | | Total |
| | Test 70% | Train 30% | | | Test 70% | Train 30% | |
|---|---|---|---|---|---|---|---|
| N225 | 283 | 120 | 403 | MC | 593 | 254 | 847 |
| IXIC | 391 | 167 | 558 | HO | 425 | 181 | 606 |
| HSI | 258 | 110 | 368 | ER | 375 | 160 | 535 |
| GSPC | 568 | 243 | 811 | FB | 425 | 181 | 606 |
| DJI[1] | 274 | 117 | 391 | MS | 492 | 210 | 702 |
| DJI[2] | 1,189 | 509 | 1,698 | TR | 593 | 254 | 847 |

*[1]. Monthly; [2]. Weekly*

### 4.3 Assessment Metric

The Root-Mean-Square Error (RMSE) is a measure frequently used for assessing the accuracy of prediction obtained by a model. It measures the differences or residuals between actual and predicated values. The metric compares prediction errors of different models for a particular data and not between datasets. The formula for computing RMSE is as follows:

---

[1] https://finance.yahoo.com/
[2] https://fred.stlouisfed.org/
[3] http://www.imf.org/external/index.htm

$$RMSE = \sqrt{\frac{1}{N}\sum_{i=1}^{N}(x_i - \hat{x}_i)^2}$$

Where N is the total number of observations. $x_i$ is the actual value; whereas, $\hat{x}_i$ is the predicated value. The main benefit of using RMSE is that it penalizes large errors. It also scales the scores in the same units as the forecast values (i.e., per month for this study).

## 5 Algorithms

The ARIMA and LSTM algorithms developed for forecasting the time series are based on "*Rolling Forecasting Origin*" [Hyndman and Athanaspoulos, 2014]. The rolling forecasting origin focuses on a single forecast, i.e., the next data point to predict, for each data set. This approach uses training sets, each one containing one more observation than the previous one, one-month look-ahead view of the data. There are several variations of rolling forecast [Hyndman, 2014]:

- **One-step forecasts without re-estimation**. The model estimates a single set of training data and then one-step forecasts are computed on the remaining data sets.
- **Multi-step forecasts without re-estimation**. Similar to one-step forecasts when performed for the next multiple steps.
- **Multi-step forecasts with re-estimation**. An alternative approach where the model is re-fitted at each iteration before each forecast is performed.

With respect to our data sets, where there is a dependency in prior time steps, a rolling forecast is required. An intuitive and basic way to perform the rolling forecast is to re-build the model when each time a new observation is added. A rolling forecast is sometimes called "*walk-forward model validation*." Python was used for implementing the algorithms along with Keras, an open source neural network library, and Theano, a numerical computation library, both for Python. The experiments were executed on a cluster of high performance computing facilities.

### 5.1 The ARIMA Algorithm

ARIMA is a class of models that captures temporal structures in time series data. ARIMA is a linear regression-based forecasting approach. Therefore it is best for forecasting one-step out-of-sample forecast. Here, the algorithm developed performs multi-step out-of-sample forecast with re-estimation, i.e., each time the model is re-fitted to build the best estimation model [Brownlee, 2017]. The algorithm, listed in Box 1, takes as input "time series" data set, builds a forecast model and reports the root mean-square error of the prediction. The algorithm first splits the given data set into train and test sets, 70% and 30%, respectively (Lines 1-3). It then builds two data structures to hold the accumulatively added training data set at each iteration, "*history*", and the continuously predicted values for the test data sets, "*prediction*."

As mentioned earlier, a well-known notation typically used in building an ARIMA model is $ARIMA\,(p,d,q)$, where:

- $p$ is the number of lag observations utilized in training the model (i.e., lag order).
- $d$ is the number of times differencing is applied (i.e., degree of differencing).
- $q$ is known as the size of the moving average window (i.e., order of moving average).

Through Lines 6-12, first the algorithm fits an $ARIMA\,(5,1,0)$ model to the test data (Lines 7-8). A value of 0 indicates that the element is not used when fitting the model. More specifically, an $ARIMA\,(5,1,0)$ indicates that the lag value is set to 5 for autoregression. It uses a difference order of 1 to make the time series stationary, and finally does not consider any moving average window (i.e., a window with zero size). An $ARIMA\,(5,1,0)$ forecast model is used as the baseline to model the forecast. This may not be the optimal model, but it is generally a good baseline to build a model, as our explanatory experiments indicated.

The algorithm then forecast the expected value (*hat*) (Line 9), adds the hat to the prediction data structure (Line 10), and then adds the actual value to the test set for refining and re-fitting the model (Line 12). Finally, having built the prediction and history data structures, the algorithm calculates the RMSE values, the performance metric to assess the accuracy of the prediction and evaluate the forecasts (Lines 14-15).

**Box 1. The developed rolling ARIMA algorithm.**

```
# Rolling ARIMA
Inputs: series
Outputs: RMSE of the forecasted data
#  Split data into 70% training and 30% testing data
1. size ← length(series) * 0.70
2. train ← series[0…size]
3. test ← series[size…length(size)]
#  Data structure preparation
4. history ← train
5. predictions ← empty
# Forecast
6. for each t in range(length(test)) do
7.     model ← ARIMA(history, order=(5, 1, 0))
8.     model fit ← model.fit()
9.     hat ← model_fit.forecast()
10.    predictions.append(hat)
11.    observed ← test[t]
12.    history.append(observed)
13. end for
14. MSE = mean_squared_error(test, predictions)
15. RMSE = sqrt(MSE)
Return RMSE
```

## 5.2 The LSTM Algorithm

Unlike modeling using regressions, in time series datasets there is a sequence of dependence among the input variables. Recurrent Neural Networks are very powerful in handling the dependency among the input variables. LSTM is a type of Recurrent Neural Network (RNN) that can hold and learn from long sequence of observations. The algorithm developed is a multi- step univariate forecast algorithm [Brownlee, 2016].

**Box 2. The developed rolling LSTM algorithm.**

```
# Rolling LSTM
Inputs: series
Outputs: RMSE of the forecasted data
# Split data into 70% training and 30% testing data
1. size ← length(series) * 0.70
2. train ← series[0…size]
3. test ← series[size…length(size)]

# Set the random seed to a fixed value for replication purpose
4. set random.seed(7)

# Fit an LSTM model to training data
Procedure fit_lstm(train, epoch, neurons)
5.    X ← train
6.    y ← train - X
7.    model = Sequential()
8.    model.add(LSTM(neurons), stateful=True))
9.    model.compile(loss='mean_squared_error', optimizer='adam')
10.   for each i in range(epoch) do
11.      model.fit(X, y, epochs=1, shuffle=False)
12.      model.reset_states()
13. end for
return model

# Make a one-step forecast
Procedure forecast_lstm(model, X)
14. yhat ← model.predict(X)
return yhat

15. epoch ← 1
16. neurons ← 4
17. predictions ← empty

# Fit the lstm model
18. lstm_model = fit_lstm(train, epoch, neurons)

# Forecast the training dataset
19. lstm_model.predict(train)

# Walk-forward validation on the test data
20. for each i in range(length(test)) do
21.    # make one-step forecast
22.    X ← test[i]
23.    yhat ← forecast_lstm(lstm_model, X)
24.    # record forecast
25.    predictions.append(yhat)
26.    expected ← test[i]
27. end for

28. MSE ← mean_squared_error(expected, predictions))
29. RMSE ← sqrt(MSE)
Return RMSE
```

To implement the algorithm, Keras library along with Theano were installed on a cluster of high performance computing center. The LSTM algorithm developed is listed in Box 2.

To be consistent with the ARIMA algorithm and in order to have a fair comparison, the algorithm starts with splitting the dataset into 70% training and 30% testing, respectively (Lines 1-3). To ensure the reproduction of the results and replications, it is advised to fix the random number see. In Line 4, the seed number is fixed to 7.

The algorithm defines a function called "*fit_lstm*" that trains and builds the LSTM model. The function takes the training dataset, the number of epochs, i.e., the number of time a given dataset is fitted to the model, and the number of neurons, i.e., the number of memory units or blocks. Line 8 creates an LSTM hidden later. As soon as the network is built, it must be compiled and parsed to comply with the mathematical notations and conventions used in Theano. When compiling a model, a loss function along with an optimization algorithm must be specified (Line 9). The "mean squared error" and "ADAM" are used as the loss function and the optimization algorithm, respectively.

After compilation, it is time to fit the model to the training dataset. Since the network model is stateful, the resetting stage of the network must be managed and controlled specially when there is more than one epoch (Lines 10 - 13). Furthermore, since the objective is to train an optimized model using earlier stages, it is necessary to set the shuffling parameter to false in order to improve the learning mechanism. In Line 12, the algorithm resets the internal state of the training and makes it ready for the next iteration, i.e., epoch.

A small function is created in Line 14 to call the LSTM model and predict the next step (one single look-ahead estimation) in the dataset. The number of epochs and the number of neurons are set in Lines 15-16 to 1 and 4, respectively. The operational part of the algorithm starts from Line 18 where an LSTM model is built with given training dataset, number of epoch and neurons. Furthermore, in Line 19 the forecast is taking place for the training data. Lines 20 - 27 use the built LSTM model to forecast the test dataset, and Lines 28 - 29 report the obtained RMSE values. It is important to note that, for reducing the complexity of the algorithm, some parts of the algorithms are not shown in Box 2 such as dense, batch size, transformation, etc. However, these parts are integral parts of the developed algorithm.

## 6   Results

The results are reported in Table 2. The data related to the financial time series or stock market show that the average Rooted Mean Squared Error (RMSE) using Rolling ARIMA and Rolling LSTM models are 511.481 and 64.213, respectively, yielding an average of 87.445 reductions in error rates achieved by LSTM. On the other hand, the economic related data show a reduction of 84.394 in RMSE where the average RMSE values for Rolling ARIMA and Rolling LSTM are computed as 5.999 and 0.936, respectively. The RMSE values clearly indicate that LSTM-based models outperform ARIMA-based models with a high margin (i.e., between 84% - 87% reduction in error rates).

## 7   Discussion: The Impact of the Number of Iterations on Fitting Models

The remarkable performance observed through deep learning-based approaches to the prediction problem is mostly due to the "*iterative*" optimization algorithm used in these approaches with the goal of finding the best results. By iterative we mean obtain the results several times and then select the most optimal one, i.e., the iteration that minimizes the errors. As a result, the iterations help in an under-fitted model to be transformed to a model optimally fitted to the data.

Table 2. The RMSEs of ARIMA and LSTM models.

| Financial Data | RMSE | | % Reduction in RMSE | Economic Data | RMSE | | % Reduction in RMSE |
| --- | --- | --- | --- | --- | --- | --- | --- |
| | ARIMA | LSTM | | | ARIMA | LSTM | |
| N225 | 766.45 | 105.315 | -86.259 | MC | 0.81 | 0.801 | -1.111 |
| IXIC | 135.607 | 22.211 | -83.621 | HO | 0.522 | 0.43 | -17.624 |
| HSI | 1,306.954 | 141.686 | -89.159 | ER | 1.286 | 0.251 | -80.482 |
| GSPC | 55.3 | 7.814 | -85.869 | FB | 0.478 | 0.397 | -16.945 |
| DJI[1] | 516.979 | 77.643 | 84.981 | MS | 30.231 | 3.17 | -89.514 |
| DJI[2] | 287.6 | 30.61 | -89.356 | TR | 2.672 | 0.569 | -78.705 |
| *Average* | *511.481* | *64.213* | *-87.445* | *Average* | *5.999* | *0.936* | *-84.394* |

[1]. Monthly; [2]. Weekly

The iterative optimization algorithms in deep learning often works around tuning model parameters with respect to the "***gradient descent***" where i) gradient means the rate of inclination of a slope, and ii) descent means the instance of descending. The gradient descent is associated with a parameter called "***learning rate***" that represents how fast or slow the components of the neural network are trained. There are several reasons that might explain why training a network would need more iterations than training other networks. The more compelling reason is the case when the data size is too big and thus it is practically infeasible to pass all the data to the training model at once. An intuitive solution to overcome this problem is to divide the data into smaller sizes and then train the model with smaller chunks one by one (i.e., batch size and iteration). Accordingly, the weights of the parameters of the neural networks are updated at the end of each step to account for the variance of the newly added training data. It is also possible that we train a network with the same training dataset more than once in order to optimize the parameters (i.e., epoch).

## 7.1 Batch Size and Iterations

If data is too big to be given to a neural network for training purposes at once, we may need to divide it into several smaller batches and train the network in multiple stages. The batch size refers to the total number of training data used. In other words, when a big dataset is divided into smaller number of chunks, each chunk is called a batch. On the other hand, iteration is the number of batches needed to complete training a model using the entire dataset. In fact, the number of batches is equal to number of iterations for one round of training using the entire set of data. For instance, assume that we have 1000 training examples. We can divide 1000 examples into batches of 250, which means four iterations would be needed to use the entire dataset for one round of training.

### 7.2 Epoch

Epoch represents the total number of times a given dataset is used for training purposes. More specifically, one epoch indicates that an entire dataset is given to a model only *once*, i.e., the dataset is passed forward and then backward through the network only *once*. Since deep learning algorithms use gradient descent to optimize their models, it makes sense to pass the entire dataset through a single network multiple times with the goal of updating the weights and thus obtaining a better and more accurate prediction model. However, it is not clear how many rounds, i.e., epoch, would be required to train a model with the same dataset in order to obtain the optimal weights. Different datasets exhibit different behavior and thus different epoch might be needed to optimally train their networks.

### 7.3 The Impact of Epoch

The study of the influence of the number of training rounds (epochs) on the same data is the focus of this section. The authors performed a series of experiments and sensitivity analysis in which they controlled the values of epoch and captured the error rates. The epoch values varied between 1-100 for each dataset. The reason for choosing 100 as the upper level value for epochs was practical and feasibility of the experiments. Computing error rates for epoch values between 1-100 took almost eight hours CPU time on a high performance-computing cluster with Linux operating system. A pilot study on the data set with Epoch = 500 ran for one week on the cluster without any outputs. The results of the sensitivity analysis and experiments are depicted in Figures 1 and 2 for financial and economic time series, respectively.

As demonstrated in Figures 1 and 2, there is no evidence that training the network with same dataset more than once would improve the accuracy of the prediction. In some cases, the performance even gets worse indicating that the trained models are being over-fitted. However, as a take away lesson, it is apparent that setting epoch = 1 does generate a reasonable prediction model and thus there is no need for further training on the same data.

## 8 Conclusion

With the recent advancement on developing sophisticated machine learning-based techniques and in particular deep learning algorithms, these techniques are gaining popularities among researchers across divers disciplines. The major question is then how accurate and powerful these newly introduced approaches are when compared with traditional methods. This paper compares the accuracy of ARIMA and LSTM, as representative techniques when forecasting time series data. These two techniques were implemented and applied on a set of financial data and the results indicated that LSTM was superior to ARIMA. More specifically, the LSTM-based algorithm improved the prediction by 85% on average compared to ARIMA. Furthermore, the paper reports no improvement when the number of epochs is changed.

The work described in this paper advocates the benefits of applying deep learning-based algorithms and techniques to the economics and financial data. There are several other prediction problems in finance and economics that can be formulated using deep learning. The authors plan to investigate the improvement achieved through deep learning by applying these techniques to some other problems and datasets with various numbers of features.

**Figure 1. The influence of the number of epochs: Financial time series data.**

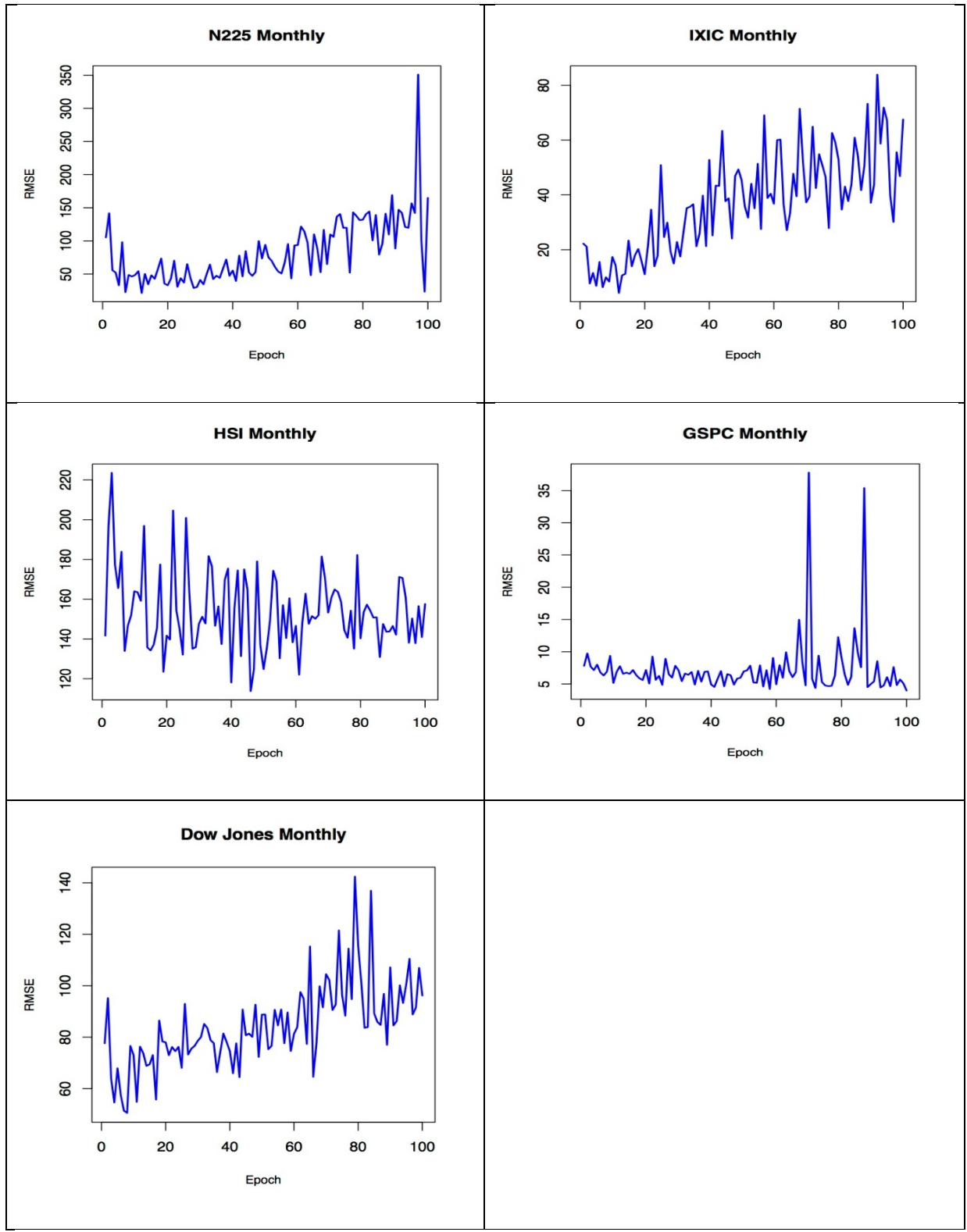

**Figure 2. The influence of the number of epochs: economic time series data.**

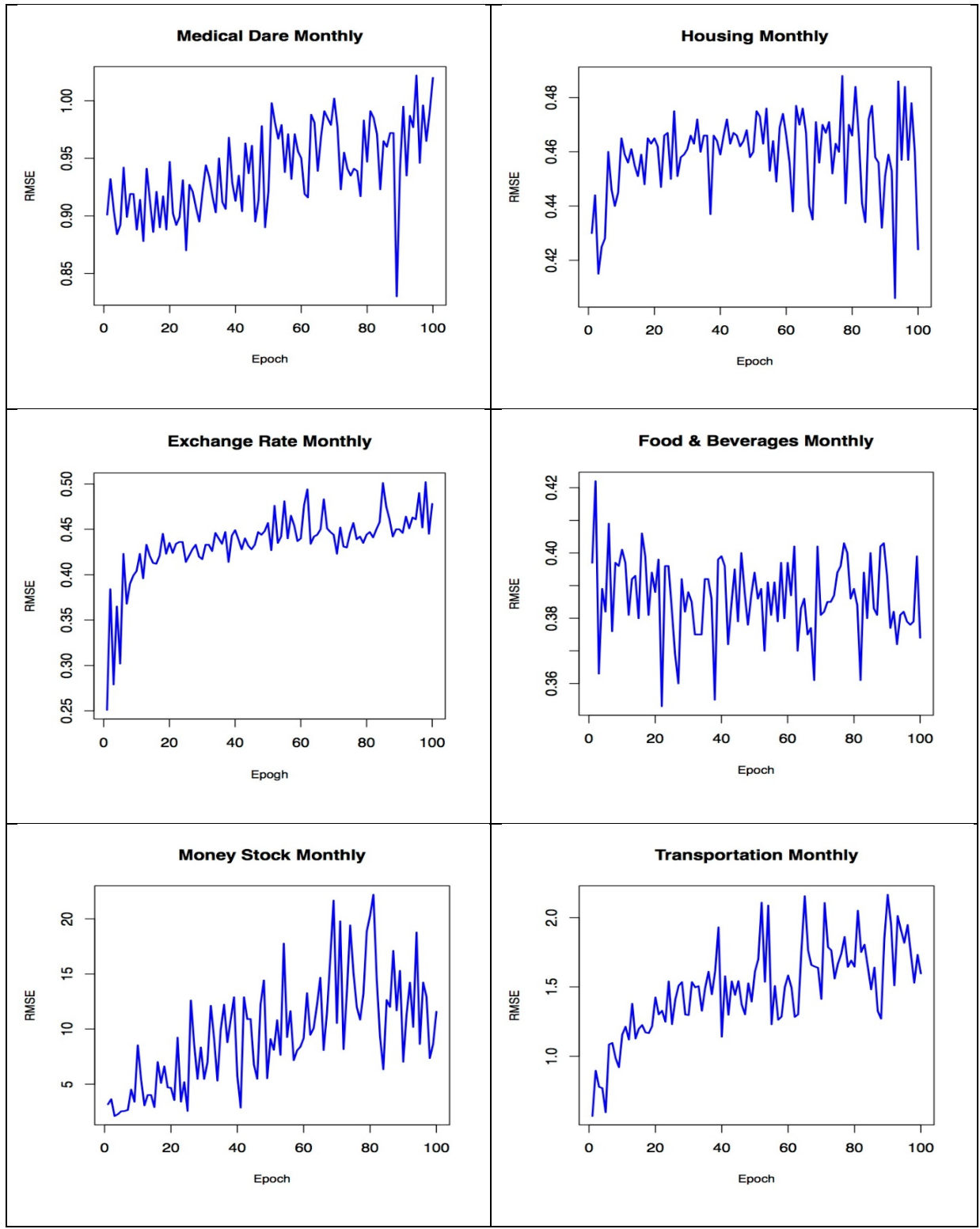

# References

Adebiyi A.A., Adewumi A.O., Ayo C.K., 2014. Stock Price Prediction Using the ARIMA Model. UKSim-AMSS 16th International Conference on Computer Modelling and Simulation.

Alonso A.M., García-Martos C., 2012. Time Series Analysis - Forecasting with ARIMA models. Universidad Carlos III de Madrid, Universidad Politécnica de Madrid.

Armstrong J.S., 2001. Evaluating Forecasting Methods. In Principles of Forecasting. A Handbook for Researchers and Practitioners (Ed. J. Scott Armstrong). Kluwer.

Banerjee A., Marcellino M., Masten I. 2005. Forecasting Macroeconomic Variables for the New Member States of the European Union. European Central Bank. Working paper Series 482.

Brownlee J., 2017. How to Create an ARIMA Model for Time Series Forecasting with Python, https://machinelearningmastery.com/arima-for-time-series-forecasting-with-python/

Brownlee J., 2016. Time Series Prediction with LSTM Recurrent Neural Networks in Python with Keras. https://machinelearningmastery.com/time-series-prediction-lstm-recurrent-neural-networks-python-keras/

Box G., Jenkins G. (1970). Time Series Analysis: Forecasting and Control. San Francisco: Holden-Day.

Colah's Blog, 2015. Understanding LSTM Networks. http://colah.github.io/posts/2015-08-Understanding-LSTMs/

Ding X., Zhang Y., Liu T., Duan J., 2015. Deep Learning for Event-Driven Stock Prediction. Proceedings of the Twenty-Fourth International Joint Conference on Artificial Intelligence (IJCAI): 2327-2333.

Earnest A., Chen M.I., Ng D., Sin L.Y. 2005. Using Autoregressive Integrated Moving Average (ARIMA) Models to Predict and Monitor the Number of Beds Occupied During a SARS Outbreak in a Tertiary Hospital in Singapore. BMC Health Service Research 5(36).

Eyben F., Wollmer M., Schuller B., Graves A., 2009. From Speech to Letters Using a Novel Neural Network Architecture for Grapheme Based ASR. in Automatic Speech Recognition & Understanding (ASRU). IEEE Workshop on. IEEE: 376-380.

Fischera T., Kraussb C., 2017. Deep Learning with Long Short-term Memory Networks for Financial Market Predictions. FAU Discussion Papers in Economics 11.

Graves A., Liwicki M., Fernandez S., Bertolami R., Bunke H., Schmidhuber J., 2009. A Novel Connectionist System for Improved Unconstrained Handwriting Recognition. IEEE Transactions on Pattern Analysis and Machine Intelligence 31(5).

Gamboa J., 2017. Deep Learning for Time-Series Analysis. University of Kaiserslautern, Kaiserslautern, Germany.

Graves A., Jaitly N., Mohamed A., 2013. Hybrid Speech Recognition with Deep Bidirectional LSTM. IEEE Workshop on In Automatic Speech Recognition and Understanding (ASRU). pp. 273-278.

Gers F.A., Schmidhuber J., Cummins F., 2000. Learning to Forget: Continual Prediction with LSTM. Neural Computation 12(10): 2451-2471.

Hochreiter S., Schmidhuber J. 1997. Long Short-Term Memory. *Neural Computation* 9(8):1735-1780.

Huck N., 2009. Pairs Selection and Outranking: An Application to the S&P 100 Index. *European Journal of Operational Research* 196(2): 819-825.

Hyndman R.J., Athanasopoulos G. 2014. Forecasting: Principles and Practice. OTexts.

Hyndman R.J. 2014. Variations on Rolling Forecasts. https://robjhyndman.com/hyndsight/rolling-forecasts/

Joshi K., Bharathi H.N., Rao J. 2013. Stock Trend Prediction using News Sentiment Analysis. https://arxiv.org/ftp/arxiv/papers/1607/1607.01958.pdf

Kane M.J., Price N., Scotch M., Rabinowitz P., 2014. Comparison of ARIMA and Random Forest Time Series Models for Prediction of Avian Influenza H5N1 Outbreaks. BMC Bioinformatics 15(1), 276.

Khashei M., Bijari M. 2011. A Novel Hybridization of Artificial Neural Networks and ARIMA Models for Time Series forecasting. *Applied Soft Computing* 11(2): 2664-2675.

Kohzadi N., Boyd M. S., Kermanshahi B., 1996. A Comparison of Artificial Neural Network and Time Series Models for Forecasting Commodity Prices. *International Journal of Neuro-computing 10: 169-181.*

Lee S.I., Seong Joon Yoo S.J., 2017. A Deep Efficient Frontier Method for Optimal Investments. Department of Computer Engineering, Sejong University, Seoul, 05006, Republic of Korea.

Patterson J., 2017. Deep Learning: A Practitioner's Approach, O'Reilly Media.

Pesaran H., 2015. Time Series and Panel Data Econometrics, Oxford University Press.

Robinson A.J., Cook G.D., Ellis D.P.W., Fosler-Lussier E., Renals S.J., Williams D.A.G., 2002. Connectionist Speech Recognition of Broad Cast News. Speech Communications. 37(1-2): 27-45.

Roondiwala M. Patel H., Varma S., 2017. Predicting Stock Prices Using LSTM. *International Journal of Science and Research (IJSR)* 6(4).

Sak H., Senior A., Beaufays F., 2014. Long Short-Term Memory Based Recurrent Neural Network Architectures for Large Vocabulary Speech Recognition. ArXiv E-prints.

Sak H., Senior A., Beaufays F. 2014. Long Short-Term Memory Recurrent Neural Network Architectures for Large Scale Acoustic Modeling. Google, USA.

Schmidhuber J., 2015. Deep learning in neural networks: An overview. Neural Networks 61: 85-117.

Siah K. W., Myers, P.D., Stock Market Prediction Through Technical and Public Sentiment Analysis. http://kienwei.mit.edu/sites/default/files/images/stock-market-prediction.pdf

Xiong R., Eric P. Nichols E.P.,, Yuan Shen Y., 2015. Deep Learning Stock Volatility with Domestic Trends. https://arxiv.org/abs/1512.04916

Yahoo Finance, Business Finance, Stock Market, Quotes, News, https://finance.yahoo.com/

Yim J. 2002. A Comparison of Neural Networks with Time Series Models for Forecasting Returns on Stock Market Index. Springer-Verlag Berlin Heidelberg, IEA/AIE, LNAI 2358: 25-35.

Zhang P.G., Qi M., 2005. Neural Network Forecasting for Seasonal and Trend Time Series, European *Journal of Operational Research* 160: 501-514.